\documentclass[journal]{IEEEtran}
 
\usepackage{graphicx}
\usepackage{booktabs}
\usepackage{amsmath}
\usepackage{url}

\begin{document}

\title{Hybrid Quantum Neural Network in High-dimensional Data Classification}
\author{*1. Hao-Yuan Chen, 2. Yen-Jui Chang, 3. Shih-Wei Liao, 3. 4. Ching-Ray Chang
\\1 Dept of Computer Science, Goldsmiths, University of London, London WC1E 7HU, United Kingdom
\\2 Dept of Physics, National Taiwan University, Taipei 106216, Taiwan 
\\3 Dept of Computer Science and Information Engineering, National Taiwan University, Taipei 106216, Taiwan 
\\4 Quantum Information Center, Chung Yuan Christian University, Taoyuan 320314, Taiwan
\\ *E-mail: (hc118@student.london.ac.uk)
\thanks{Manuscript}}

\maketitle

\begin{abstract}
The research explores the potential of quantum deep learning models to address challenging machine learning problems that classical deep learning models find difficult to tackle. We introduce a novel model architecture that combines classical convolutional layers with a quantum neural network, aiming to surpass state-of-the-art accuracy while maintaining a compact model size. The experiment is to classify high-dimensional audio data from the Bird-CLEF 2021 dataset. Our evaluation focuses on key metrics, including training duration, model accuracy, and total model size. This research demonstrates the promising potential of quantum machine learning in enhancing machine learning tasks and solving practical machine learning challenges available today.
\end{abstract}

\begin{IEEEkeywords}
Quantum deep learning utility, Hybrid Quantum neural network, High-dimensional data, Binary classification
\end{IEEEkeywords}

\IEEEpeerreviewmaketitle

\section{Introduction}
\IEEEPARstart{Q}{uantum} Machine learning is a promising research domain where quantum computing and machine learning are combined to solve various challenges in machine learning problems and other applicable challenges in quantum computing. High-dimensionality of the quantum information implies a significant opportunity for machine learning and deep learning models to be improved and accelerated by quantum computing techniques. In the research, \cite{Biamonte2017} has illustrated the overall frameworks and promising acceleration of various quantum computing and quantum-inspired computations in machine learning domains. The research aims to address whether there is any hybrid quantum neural network that could deliver a significant advancement of classical deep learning models from the perspectives of accuracy, model size, and training speed in a binary classification problem under a high-dimensional dataset.

The hypothesis, in addition, is based on the concept that the hybrid quantum deep neural network could provide a significant advancement in the area where classical neural networks struggle to deliver excellent results in which the SOTA model can’t correctly have an accuracy of more than 90 percent. Moreover, the benefits of the hybrid quantum neural network consist of accuracy improvement and model size reduction. The selection dataset in this study is challenging yet crucial for environmental and biological science research. However, to the best of our knowledge, by 2023, the SOTA model of the Bird-CLEF 2021 can’t surpass the threshold of 90 percent accuracy with results at around 88 percent prediction accuracy. 

The research objectives of this research are, firstly, designing a hybrid quantum neural network on top of the current architecture of a state-of-the-art (SOTA) model combining a quantum neural network with classical optimizers like the Adam optimization algorithm. The second goal is to experiment and validate the hybrid neural network against the classical SOTA models to ensure it surpasses the overall performance benchmark from various metrics. At the end of the study, a discussion on the potential quantum deep learning utility is delivered to evaluate the general utility of the hybrid quantum model with the metrics from classification accuracy, total model size, and the training speed (represented in training duration in seconds).

\section{Binary Classification in High-dimensional Datasets}
Firstly, a theoretical framework for this research is constructed by discussing binary classification using a convolutional neural network and the challenges of high-dimensional datasets in machine learning. The theoretical framework of this research provides an analytical view to model and understand the problem we are approaching to deliver quantum deep learning utility with a novel hybrid quantum deep learning model.

\subsection{Binary Classifications using Convolutional Neural Network (CNN)}
Binary classification is a commonly seen classification and machine learning problem in which algorithms aim to correctly identify the input data by binary classes. The mathematical model that can describe binary classification generally revolves around finding a function that maps input features to a binary output (0 or 1, representing the two classes). 

In binary classification with Convolutional Neural Networks (CNNs), the architecture learns features from images through layers like convolutional layers with filters for feature mapping and ReLU activation functions for non-linearity. Pooling layers reduce feature map dimensions, enhancing efficiency. The network then synthesizes these features in fully connected layers for high-level reasoning. This leads to a sigmoid-activated output layer that gives a probability score for classifying inputs into two categories. The network is optimized using algorithms like Stochastic Gradient Descent and backpropagation, with binary cross-entropy as the loss function, enabling effective image-based binary classification.

\subsection{High Dimensional Datasets}
High-dimensional datasets, characterized by many features relative to the number of observations, present significant challenges for machine learning models, a phenomenon known as the "curse of dimensionality." In such environments, models are prone to overfitting, where they excel at memorizing the training data, including noise, but fail to generalize to new, unseen data. The sparsity of data in high-dimensional spaces complicates the learning process further, as the volume of data required to represent the space adequately grows exponentially with each added dimension. Traditional distance metrics, crucial for many algorithms, lose their effectiveness, as points in high-dimensional spaces tend to be equidistant from each other, obscuring meaningful relationships. 

Quantum algorithms, characterized by their potential for exponential speed-ups in certain computations, can accelerate critical machine learning tasks like optimization and searching. They are particularly adept at managing sparse datasets, a common feature in high-dimensional scenarios. Moreover, quantum approaches to feature selection and dimensionality reduction could outperform classical methods, more effectively identifying the most relevant features. Enhanced calculation of distance metrics, crucial in many classification algorithms, is another area where quantum computing could substantially improve.

The dataset used in this research, Bird-CLEF 2021, officially known as the "Bird Call Identification Challenge 2021," was a notable event in bioacoustics and artificial intelligence. Hosted on the renowned platform CLEF (Conference and Labs of the Evaluation Forum), this challenge focused on identifying bird species through their unique vocalizations. The competition drew attention from diverse participants, including data scientists, ecologists, and bird enthusiasts. By leveraging advanced machine learning algorithms and extensive datasets of bird calls, participants aimed to develop models that could accurately recognize and classify bird species based on their calls. The outcomes of Bird-CLEF 2021 were expected to have significant implications in the fields of ornithology and conservation biology, showcasing the potential of AI in understanding and preserving natural ecosystems.

\section{Methods}
Classical machine learning, especially deep learning, has witnessed remarkable success in recent years, primarily due to its ability to model complex datasets such as images and videos effectively. However, a bottleneck has emerged in this domain, where classical deep neural networks (DNNs) face immense challenges when confronted with high-dimensional datasets, exemplified by the Bird-CLEF series, which forms the basis of this research. However, with the recent advancements in quantum machine learning, our study offers a fresh perspective on overcoming these challenges using today's general quantum hardware and classical quantum simulators.

Therefore, the study proposes a novel hybrid quantum neural network (H-QNN) by combining pre-trained convolutional layers and a quantum neural network comprised of a quantum feature map and Ansatz to perform the binary classification in a one-qubit configuration. The overall structure of the deep neural network is shown in Table I. The aim is to utilize the superposition nature of quantum mechanics to efficiently and effectively classify the input features encoded by the quantum feature maps in a variational manner. The one-qubit state could present two possible outcomes of information with distribution in probability, making an ideal binary classifier with a compact model size.

\subsection{Hybrid Model Architecture}
The predominant paradigm in quantum deep learning often revolves around using variational quantum circuits as trainable neural networks. These circuits optimize quantum gate parameters to match specific probability distributions or desired outcomes. However, given that most datasets currently in use are classical and pose challenges for direct quantum encoding, our research introduces an alternative approach. Instead of employing quantum neural networks directly, we incorporate classical convolutional layers as feature extractors. These layers facilitate the transformation of classical data into more amenable features, which are subsequently encoded into quantum states via quantum feature maps, notably the Z feature maps. 

Within the quantum neural network, we design an Ansatz to effectively learn the distribution of input features and produce binary outputs for classification using a one-qubit quantum circuit. The quantum neural network is encapsulated using IBM Qiskit’s Sampler primitive with Sampler-QNN architecture. The sampler QNN returns the distribution of the quantum circuit for all possible states making a one-qubit neural network that can present binary results with two different states of outcomes in probability. This architectural innovation has yielded substantial improvements in classification accuracy and reduced overall model size, resulting in a notable advancement. Furthermore, it has enhanced the general utility of the model by considering three key measurement metrics in this research. 

\begin{figure}
    \centering
    \includegraphics[width=1\linewidth]{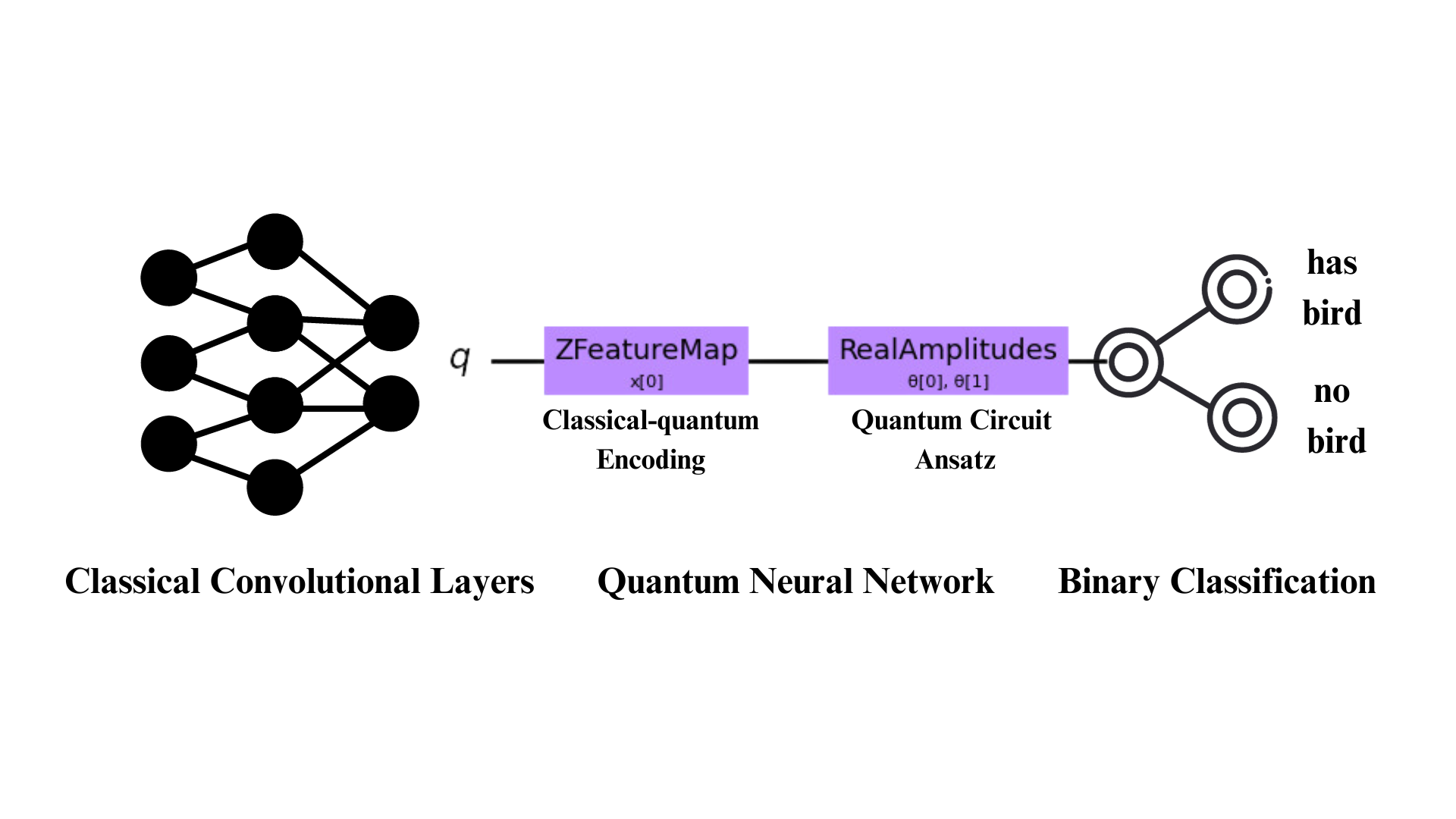}
    \caption{Hybrid Model Architecture}
\end{figure}

\begin{table}[ht]
\caption{Hybrid Model Architecture and Parameters}
\centering
\begin{tabular}{|l|l|l|}
\hline
\textbf{Layer (type)} & \textbf{Output Shape} & \textbf{Param \#} \\
\hline
Conv2d-1 & [-1, 64, 64, 64] & 9,408 \\
BatchNorm2d-2 & [-1, 64, 64, 64] & 128 \\
ReLU-3 & [-1, 64, 64, 64] & 0 \\
MaxPool2d-4 & [-1, 64, 32, 32] & 0 \\
Conv2d-5 & [-1, 128, 32, 32] & 8,192 \\
BatchNorm2d-6 & [-1, 128, 32, 32] & 256 \\
ReLU-7 & [-1, 128, 32, 32] & 0 \\
Conv2d-8 & [-1, 128, 32, 32] & 4,608 \\
BatchNorm2d-9 & [-1, 128, 32, 32] & 256 \\
... & ... & ... \\
Linear-100 & [-1, 1] & 513 \\
TorchConnector-101 & [-1, 2] & 2 \\
\hline
\textbf{Total params} & & 1,412,931 \\
\textbf{Trainable params} & & 1,412,931 \\
\textbf{Non-trainable params} & & 0 \\
\hline
\end{tabular}
\end{table}

\subsection{Pre-trained Convolutional Neural Network (CNN)}
One of the novel advancements of this research is to incorporate a pre-trained convolutional neural network  (CNN). The pre-trained model of the choice is the ResNext CNN model. However, the pre-trained CNN in this research was modified to remove the dense layers to pass the features from the convolutional layers directly to the quantum neural network (QNN). A quantum neural network could analyze larger-scale audio and image data compared to the full-quantum neural network without pre-trained CNN.

\subsection{Single-qubit Binary Classifier}
In the study, the employment of Sampler networks within IBM Qiskit forms the core of the research. These networks encapsulate the Quantum Neural Network (QNN) by utilizing the Sampler primitive, distinguishing them fundamentally from Estimator networks. While Estimator networks solely provide the expected values of a neural network, Sampler networks offer a distribution of results deriving from a quantum circuit. This novel methodology substantially reduces the quantum circuit’s size, diminishing it by a factor proportional to the square root of $n$, wherein $n$ symbolizes the number of classification classes. As sketched in Figure 2, the QNN in this study comprises two pivotal components. Initially, the QNN employs a feature map to extract classical features using convolutional layers into quantum states. Subsequently, a quantum Ansatz is leveraged to compute quantum states, thereby mapping the features with the corresponding binary results. This operation is crucial for demonstrating the presence of protected species within audio data.

In quantum computing, particularly in binary classification using a single qubit, the probability of the qubit being measured in either of its basis states---\( |0\rangle \) or \( |1\rangle \) --- is central to the classification decision. After the qubit undergoes various transformations (represented as \( | \psi'' \rangle \)), the probabilities \( P(0) \) and \( P(1) \), which represent the likelihood of measuring the qubit in the \( |0\rangle \) and \( |1\rangle \) states respectively, are computed. These probabilities are derived from the squared magnitudes of the inner products of the qubit's final state with the basis states. The classification decision is then based on these probabilities: if the probability of the \( |0\rangle \) state is greater than that of the \( |1\rangle \) state, the classifier outputs 0; otherwise, it outputs 1. This approach effectively utilizes the probabilistic nature of quantum measurements, allowing the qubit's state to determine the classification outcome in a binary classification problem.

\begin{equation}
P(0) = |\langle 0 | \psi'' \rangle|^2
\end{equation}

\begin{equation}
P(1) = |\langle 1 | \psi'' \rangle|^2
\end{equation}

\begin{equation}
\text{Binary classes} = 
\begin{cases} 
0 & \text{if } P(0) > P(1) \\
1 & \text{otherwise}
\end{cases}
\end{equation}

\subsection{Quanutm Neural Network Design}
In the study, the employment of Sampler networks within IBM Qiskit forms the core of the research. These networks encapsulate the Quantum Neural Network (QNN) by utilizing the Sampler primitive, distinguishing them fundamentally from Estimator networks. While Estimator networks solely provide the expected values of a neural network, Sampler networks offer a distribution of results deriving from a quantum circuit.

This novel methodology substantially reduces the quantum circuit’s size, diminishing it by a factor proportional to the square root of $n$, wherein $n$ symbolizes the number of classification classes. As sketched in Figure 2, the QNN in this study comprises two pivotal components. Initially, the QNN employs a feature map to extract classical features using convolutional layers into quantum states. Subsequently, a quantum Ansatz is leveraged to compute quantum states, thereby mapping the features with the corresponding binary results. This operation is crucial for demonstrating the presence of protected species within audio data.

\subsubsection{Quantum Feature Map}
The Z-feature map in Qiskit is a critical component used in quantum machine learning and quantum computing applications. It's designed to encode classical data into a quantum state, essential for processing information using quantum algorithms. This feature map applies a series of quantum gates, typically Pauli-Z rotations, to qubits in a quantum circuit. These rotations are parameterized by the input data, effectively embedding the classical data into the quantum state of the qubits. The Z-feature map is particularly effective in constructing quantum kernels for machine learning, as it can create complex, high-dimensional quantum states that can capture intricate patterns in the data. By leveraging the principles of superposition and entanglement, the Z-feature map allows quantum computers to process and analyze data in ways that are fundamentally different from classical computers, potentially leading to more efficient and robust machine learning models.

\subsubsection{Quantum Ansatz}
The Real Amplitude ansatz is a specific type of quantum circuit design chosen for its simplicity and efficiency in representing quantum states with real amplitudes. It consists of a series of parameterized single-qubit rotational gates and entangling gates. The single-qubit gates typically include rotations around the Y axes (RY gate), parameterized by angles adjusted during optimization. The entangling gates, often CNOT gates, create quantum correlations between qubits.

In variational algorithms, the Real Amplitude ansatz is the computational unit for the quantum state. The parameters of this ansatz are iteratively optimized to minimize a cost function, which is typically related to the problem being solved, such as finding the ground state energy of a molecule in VQE or solving optimization problems in QAOA. This ansatz is favored for its ability to efficiently explore the state space with a relatively simple and shallow quantum circuit, making it suitable for near-term quantum devices that are limited in terms of qubit count and coherence times.

\begin{figure}
    \centering
    \includegraphics[width=1\linewidth]{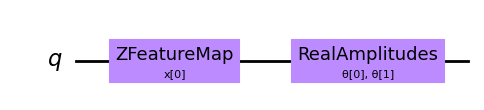}
    \caption{Hybrid Model Architecture}
\end{figure}

\section{Training Algorithms}
The training algorithm presented in this research outlines a comprehensive and structured approach for training a deep learning model, specifically a custom version of the ResNext architecture. This function is part of a larger machine-learning pipeline, as indicated by its integration with configuration settings (CFG), data loaders, and logging mechanisms. Initially, the function initializes data loaders for training and validation datasets, with appropriate transformations applied. The DataLoader objects are configured with parameters such as batch size, number of workers, and memory settings, which are crucial for efficient data handling during training. 

The model, a CustomResNext, is initialized with pre-trained weights and then transferred to the specified device (i.e., GPU). The optimizer used here is Adam, a popular choice for training deep neural networks due to its efficiency in handling sparse gradients and adaptive learning rate capabilities. The function performs training and validation steps in the main training loop for each epoch. The training step involves running the train fn function with the training data loader, while the validation step involves running the valid fn function with the validation data loader. After each epoch, it logs the average losses and the elapsed time, providing insights into the training progress.

The training process of the hybrid quantum neural network involves the backpropagation of quantum neural networks connecting with the ResNeXt pre-trained CNN via TorchConnector from IBM Qiskit. The QNN, in this setup, is responsible for outputting the probabilities of a qubit being in states |0⟩ or |1⟩, determined by the amplitudes (\(\alpha\) and \( \beta \)) of its quantum state. Once the qubit state is measured, these probabilities represent the likelihood of the two classification outcomes.

The loss function of the training process is the Cross-Entropy loss function, which is particularly suited for classification tasks, as it quantifies the difference between the predicted probabilities and the actual labels. It is mathematically expressed as \( L = -[y \log(p) + (1 - y) \log(1 - p)] \), where \( y \) is the actual label and \( p \) is the predicted probability for one of the classes. In the context of your hybrid model, \( p \) would specifically be the probability of the qubit existing in the |1⟩ state after processing through the quantum circuit.

Training such a hybrid model involves fine-tuning the ResNext CNN weights and the quantum circuit parameters to minimize this Cross-Entropy loss. This necessitates backpropagation through both the classical and quantum segments of the network. However, due to their unique computational nature, quantum circuits require specialized gradient estimation techniques, such as parameter shift rules.

Another critical aspect of this loop is evaluating model performance based on accuracy and implementing a mechanism to save the model checkpoint if the current epoch yields a better score than previous epochs. This checkpointing approach ensures that the best-performing model is retained. Finally, the function loads the best model checkpoint and updates the validation folds with predictions, returning them along with the scores for each epoch. This structured approach facilitates practical training and validation and ensures the reproducibility and monitoring of the model's performance over time.

\section{Results}
The hybrid model was evaluated by three primary metrics: model accuracy, size, and training runtime. Firstly, the model accuracy reached SOTA status by surpassing the previous record of 88.6 percent, the model's highest accuracy of 90.24 percent. The experiments were conducted and evaluated on the classical machine learning hardware with Nvidia RTX3080 Ti GPU. 

Table II compares a Hybrid Quantum Neural Network (H-QNN) and a Classical Convolutional Neural Network (CNN) using the ResNeXt architecture, evaluating their performance based on four critical metrics. Notably, the Hybrid QNN demonstrates a marginally higher accuracy (89.79\%) than its classical counterpart (88.69\%), indicating its superior ability to make correct predictions even in the high-dimensional dataset. However, this comes at the cost of increased runtime, with the Hybrid QNN taking 443.48 seconds compared to the Classical CNN - ResNeXt's 422.15 seconds. A significant advantage of the Hybrid QNN is its model size, which is substantially smaller (226.46 MB) than the Classical CNN - ResNeXt (382.74 MB), suggesting a more efficient use of storage and memory resources. This indicates that despite the slightly longer runtime, the Hybrid QNN's combination of higher accuracy and reduced model size translates into a more effective and efficient overall performance.

\begin{table}[h!]
\centering
\begin{tabular}{lcc}
\toprule
\textbf{Metric} & \textbf{Hybrid QNN} & \textbf{Classical CNN - ResNeXt} \\
\midrule
Accuracy (\%) & 89.79\% & 88.69\% \\
Runtime (sec) & 443.48 & 422.15 \\
Model Size (MB) & 226.46 & 382.74 \\
\bottomrule
\end{tabular}
\vspace{0.25cm}
\caption{Comparison of Hybrid and Classical Approaches}
\label{tab:comparison}
\end{table}

\begin{figure}[h!]
    \centering
    \begin{minipage}{0.48\linewidth}
        \includegraphics[width=\linewidth]{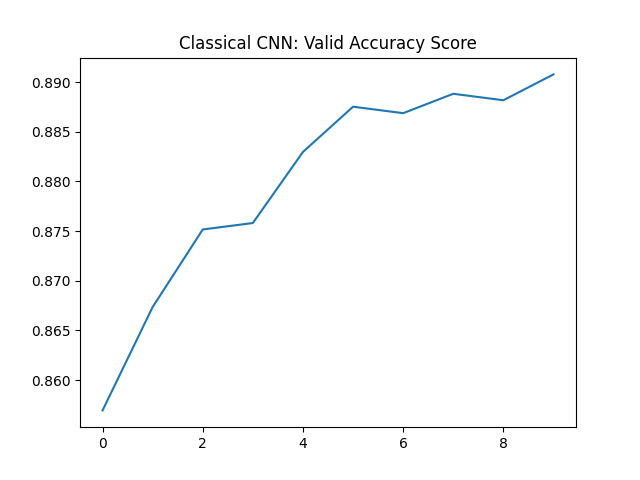}
        \caption{CNN Training Accuracy Log}
    \end{minipage}
    \hfill
    \begin{minipage}{0.48\linewidth}
        \includegraphics[width=\linewidth]{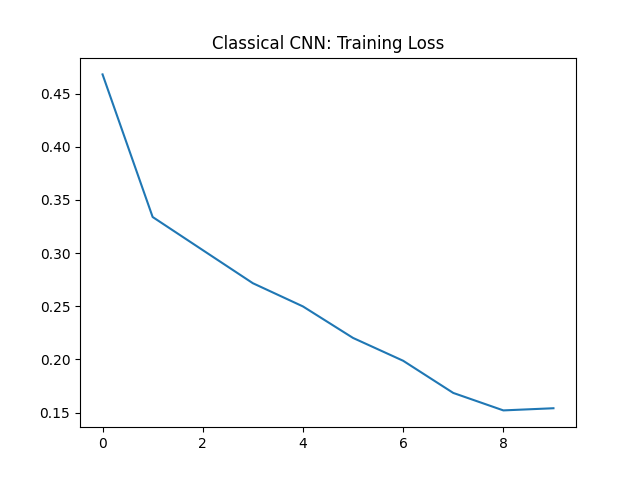}
        \caption{CNN Training Loss Log}
    \end{minipage}
\end{figure}

\begin{figure}[h!]
    \centering
    \begin{minipage}{0.48\linewidth}
        \includegraphics[width=\linewidth]{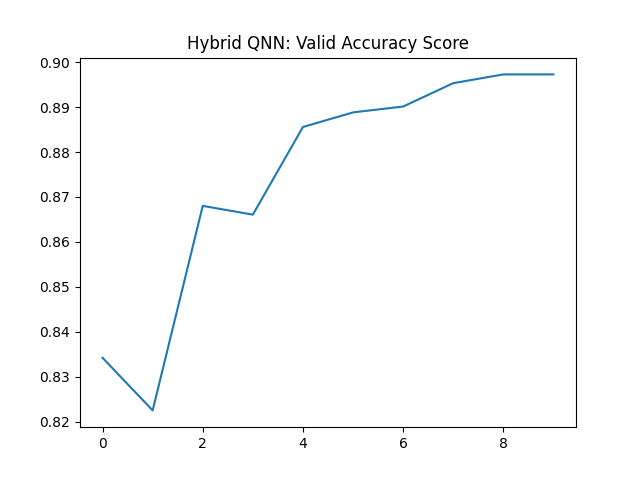}
        \caption{H-QNN Training Accuracy Log}
    \end{minipage}
    \hfill
    \begin{minipage}{0.48\linewidth}
        \includegraphics[width=\linewidth]{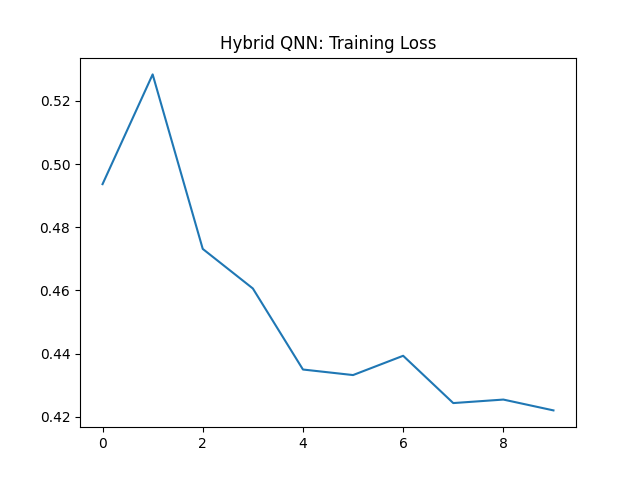}
        \caption{H-QNN Training Loss Log}
    \end{minipage}
\end{figure}

\section{Discussion}
This study introduces a novel hybrid architecture combining quantum computing and deep learning. This design features pre-trained classical convolutional layers integrated with a quantum neural network. The architecture has demonstrated promising results when applied to high-dimensional datasets, achieving enhanced prediction accuracy and reducing the model's overall size. Subsequently, the paper will present a detailed analysis of the model's advantages, limitations, and scalability. This discussion aims to provide a comprehensive understanding of the model's overall performance and outline potential future developments in this area of research.

\subsection{Advantages and Limitations}
This model's primary benefit lies in its ability to substantially enhance training speed and prediction accuracy using a one-qubit circuit. The one-qubit quantum neural network facilitates fast simulation through straightforward mapping to binary outcomes and has shown a considerable advantage in minimizing the model's size. Nonetheless, its effectiveness in addressing multiclass classification challenges is yet to be determined. Consequently, it is critical to explore whether our one-qubit architecture maintains its efficiency and effectiveness when applied to problems involving multiple classes in future research. 

\subsection{Scalability}
However, findings from this study indicate that a circuit with $n$ qubits can represent $2^n$ outcomes. Consequently, the number of classes denoted as the number of qubits $n$ can mean $c$, approximately equating to \(\sqrt{c}\). This relationship suggests a scalable approach to representing multiple qubits' quantum neural network architecture classes which presents a novel opportunity to reduce the overall model size to the larger models on multiclass classification.

\section{Summary}
The research introduced a novel quantum deep learning model combining classical convolutional layers and a quantum binary classifier built with a variational quantum circuit (VQC). The empirical results of the research suggest that this hybrid model architecture could deliver potential quantum deep learning utility from the perspectives of performance (accuracy of prediction), training runtime, and overall model size. The future outlook of this research aims to seek multi-class classification with a benchmark like ImageNet to understand the limitations and potential advances of quantum deep learning research.

\section*{Acknowledgment}
The authors thank Ph.D. candidate Yen-Jui Chang at National Taiwan University for supporting and advising this project. Also, a big thanks to Prof. Shih-Wei Liao and Prof. Ching-Ray Chang for providing professional guidance on research trajectories. We acknowledge support from the National Science and Technology Council, Taiwan, under Grants NSTC 112-2119-M-033-001, for the research project Applications of Quantum Computing in Optimization and Finances.

\section*{Declarations}
\subsection*{Competing interests}
No, I declare that the authors have no competing interests as defined by Springer or other interests that might be perceived to influence the results and discussion reported in this paper.

\subsection*{Authors' contributions}
Mr. Hao-Yuan Chen oversaw the entire research project, taking charge of various critical aspects. He skillfully designed the neural network architecture and conducted rigorous validation experiments. Furthermore, he contributed significantly to crafting most of the paper's contents. Alongside Mr. Hao-Yuan Chen, Prof. Chang and the Ph.D. candidate, Yen-Jui, were valuable collaborators. They actively engaged in stimulating and thought-provoking discussions, enriching the research process. Additionally, their efforts were focused on improving the quantum circuit design, which proved pivotal to the study's success.

\ifCLASSOPTIONcaptionsoff
  \newpage
\fi

\end{document}